\documentclass{article}
\usepackage{ijcai26}

\usepackage{times}
\usepackage{soul}
\usepackage{url}
\usepackage[hidelinks]{hyperref}
\usepackage[utf8]{inputenc}
\usepackage[small]{caption}
\usepackage{amsmath}
\usepackage{amsthm}
\usepackage{booktabs}
\usepackage{algorithm}
\usepackage{algorithmic}
\usepackage[switch]{lineno}
\usepackage{multirow}
\usepackage{graphicx}
\usepackage{subcaption}
\usepackage{amssymb}

\author{
Bohou Zhang$^{*}$
\and
Xiaoyu Tao$^{*}$\\
Mingyue Cheng$^{\dagger}$\and
Huijie Liu\and
Qi Liu
\affiliations
State Key Laboratory of Cognitive Intelligence, University of Science and Technology of China
\emails
\{zbhustc, txytiny, lhj33\}@mail.ustc.edu.cn,
\{mycheng, qiliuql\}@ustc.edu.cn
}

\title{ScholarSum: Student-Teacher Abstractive Summarization via Knowledge Graph Reasoning and Reflective Refinement}

\begin{document}

\maketitle

\renewcommand{\thefootnote}{\fnsymbol{footnote}}
\footnotetext[2]{Corresponding author: Mingyue Cheng.}
\renewcommand{\thefootnote}{\arabic{footnote}}

\begin{abstract}
Abstractive summarization plays a crucial role in enabling efficient understanding of scientific literature, yet it inherently demands both linguistic fluency and factual faithfulness. Existing approaches often fail to reconcile these two requirements. Extractive methods rely on rigid sentence splicing that disrupts macro-level logical coherence, while large language model (LLM)-based generative approaches, despite mastering linguistic fluency, exhibit limited factual consistency.
In this work, we propose ScholarSum, a hierarchical reflective graph-based framework that emulates a student–teacher writing process for fluent and faithful scientific summarization. ScholarSum first organizes the document into a hierarchical knowledge graph by segmenting it into semantically coherent units, whose multi-layered community structure captures global logic and macro-level themes. Guided by this global structure, the student generates an initial draft, which is subsequently refined through fine-grained evidence retrieval.
To ensure factual consistency, a teacher-like reviewer then iteratively examines the draft, identifies unsupported content, and prompts targeted re-retrieval and rewriting until the summary meets rigorous quality standards.
Extensive experiments demonstrate that ScholarSum significantly outperforms previous baselines in terms of both completeness and faithfulness. 
Our code is available at \url{https://github.com/Xiaoyu-Tao/ScholarSum}.

\end{abstract}

\section{Introduction}

Abstractive summarization is critical for efficient understanding of scientific literature in domains such as medicine, computer science, and the natural sciences~\cite{cohan-etal-2018-discourse,zhou2025benchmarking}. In practice, effective abstractive summarization requires not only producing linguistically fluent text, but also preserving factual faithfulness to the original scientific claims~\cite{gao2023chatgpt}. Fundamentally, abstractive summarization can be viewed as learning to reorganize and condense document-level information into a coherent and faithful textual representation~\cite{nan2021entity}.

\begin{figure}[t]
  \centering
  \includegraphics[width=\linewidth]{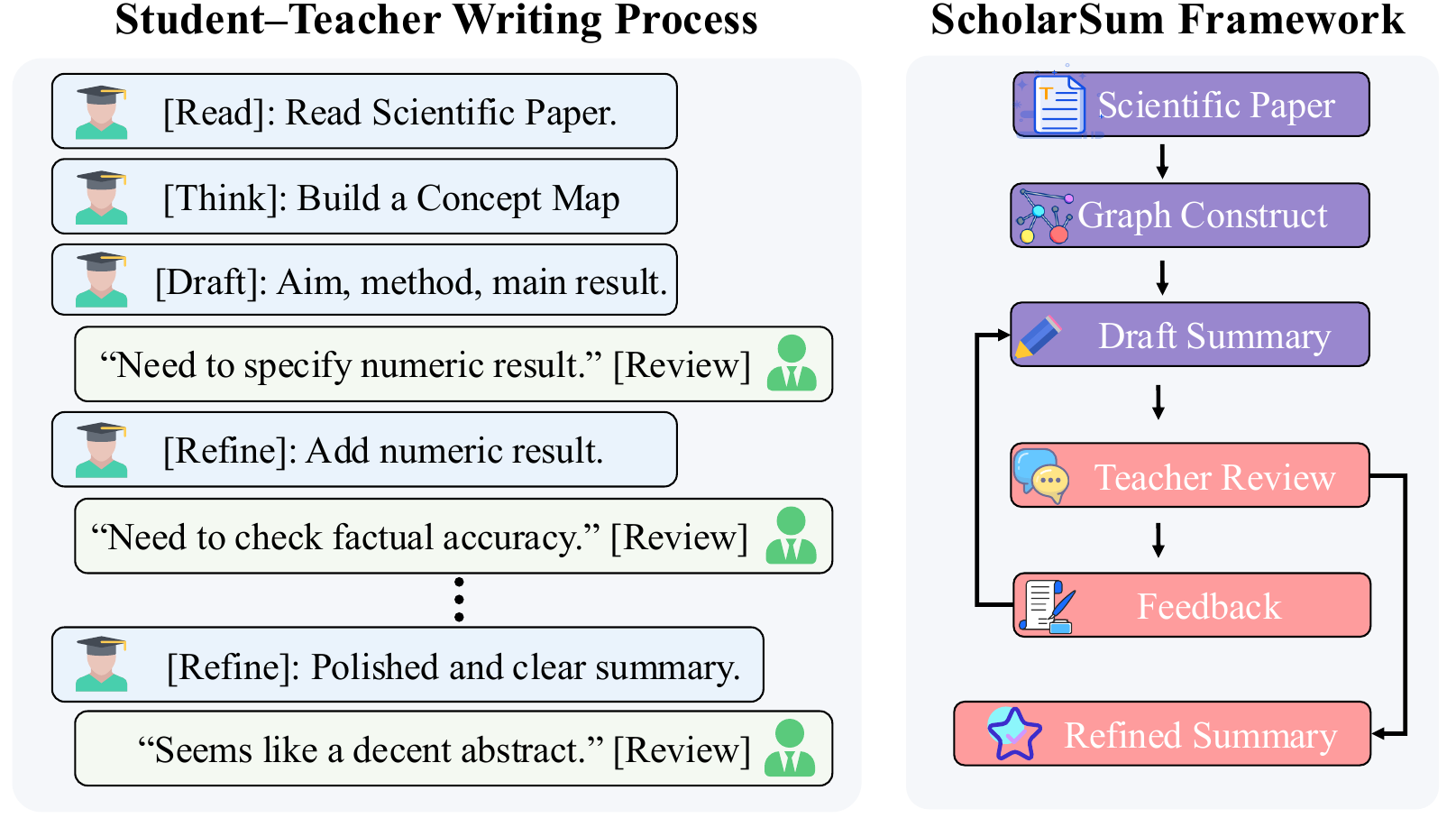}
  \caption{Illustration of how ScholarSum emulates the human student–teacher writing process using a graph-based iterative approach.}
  \label{fig:revision_comparison}
\end{figure}

The field of abstractive summarization has evolved through distinct technological developments. Early research predominantly focused on extractive methods, which relied on statistical heuristics to select representative sentences from the source text~\cite{erkan2004lexrank}. While such approaches tend to preserve factual content, their reliance on sentence selection often limits linguistic fluency. The advent of deep learning shifted the focus toward abstractive summarization, where encoder–decoder models such as BART and T5 significantly improved the fluency and expressiveness of generated text~\cite{lewis-etal-2020-bart,raffel2020exploring}. Recently, LLMs have further transformed the field by exhibiting strong zero-shot capabilities and generating coherent narratives without task-specific training~\cite{brown2020language}. However, despite these advances in linguistic fluency, ensuring factual faithfulness remains a significant challenge. This observation raises a fundamental question: \textit{Can abstractive summarization fully reconcile linguistic fluency with factual faithfulness?}

Motivated by this question, as shown in Figure~\ref{fig:revision_comparison} (left), we observe that in real-world writing practice, a student typically first reads and internalizes a paper by organizing its key ideas into a hierarchical mental representation, and then produces an initial summary that is subsequently refined through iterative feedback from a teacher. This process often leads to summaries that maintain linguistic fluency while progressively improving factual accuracy~\cite{xu2023fine}.
Inspired by this observation, as shown in Figure~\ref{fig:revision_comparison} (right), we explore an underexplored approach that formulates abstractive summarization as a hierarchical content understanding and generation problem guided by LLMs. The underlying intuition is that modeling document-level structure ensures global coherence, while iterative refinement progressively corrects missing or unsupported content.
However, this approach introduces several non-trivial challenges. First, constructing a hierarchical representation that faithfully reflects a document’s logical structure is non-trivial, especially for long and complex scientific texts~\cite{liu2023lost}. Second, even with structural guidance, effectively integrating local evidence to ensure factual precision remains difficult, as global structural coherence does not necessarily guarantee fine-grained factual grounding~\cite{sun2024prompt}.

To address these challenges, we propose ScholarSum, a hierarchical reflective graph-based framework inspired by the student–teacher writing process, designed to generate summaries that are both linguistically fluent and factually faithful. Specifically, ScholarSum first organizes the document into a hierarchical knowledge graph by segmenting it into semantically coherent units, where the multi-layered community structure corresponds to semantically coherent themes and provides a global outline for summary planning. Guided by this hierarchical structure, the student generates an initial draft, ensuring fluent expression and coherent organization. To systematically enforce factual consistency, a teacher-like reviewer then iteratively examines the draft, identifies missing or unsupported content, and prompts targeted re-retrieval and rewriting until the summary meets rigorous quality standards. This hierarchical and reflective design allows ScholarSum to fully leverage the capabilities of LLMs in semantic understanding, contextual reasoning, and coherent text generation, while explicitly addressing the dual requirements of fluency and faithfulness. Extensive experiments on benchmark scientific summarization datasets show that ScholarSum significantly outperforms previous baselines.

\section{Related Work}

In this section, we review prior work relevant to abstractive summarization, focusing on two major lines of research.

\subsection{Conventional Abstractive Summarization}
Early approaches to summarization relied heavily on extractive methods, where sentences were scored using statistical or neural metrics and top-ranked sentences were concatenated to form summaries. Although effective at capturing salient content, such methods often resulted in fragmented narratives with limited coherence, repetition, and poor stylistic fluency~\cite{erkan2004lexrank}.
Generative approaches based on pre-trained language models (PLMs) partially alleviated these issues by enabling end-to-end abstraction. Sequence-to-sequence architectures improved surface fluency, and pointer-generator mechanisms with coverage helped balance abstraction and copying~\cite{see-etal-2017-get}. Further tuning-based methods, including task-specific pretraining objectives such as gap sentence prediction and unified text-to-text frameworks, established strong baselines on scientific summarization datasets~\cite{zhang2020pegasus,raffel2020exploring}. Iterative refinement strategies, inspired by human writing processes, introduced cycles of drafting, critique, and revision~\cite{sun2024prompt,li2024isqa}. Related self-refinement frameworks also demonstrate the value of explicit feedback loops for improving factual consistency~\cite{madaan2023selfrefine}.
However, these PLM-based approaches still struggle to maintain coherent and globally consistent summaries over long documents, particularly when cross-section integration is required.

\subsection{LLM-based Summarization}
LLM-based methods leverage large-scale models to handle longer contexts and improve discourse-level fluency. One representative strategy decomposes summary generation into intermediate reasoning steps, enabling more coherent and structured outputs. Stepwise decomposition and prompt chaining outperform single-pass generation, particularly for scientific and technical texts~\cite{xu2023inheritsumm}.
Another line of work adopts retrieval-augmented generation (RAG), where LLMs retrieve key fragments or subgraphs of relevant documents prior to reasoning~\cite{asai2024selfrag,sarthi2024raptor}. Graph-aware retrieval further exploits document structure, such as citation and discourse relations, to support multi-hop reasoning and improve global coherence~\cite{edge2024local,peng2024graph}. Related graph-based planning and retrieval methods further highlight the benefits of structured evidence organization~\cite{hu-etal-2025-grag,chen2024plan}. Recent studies on scientific paper mining suggest that explicit search and tool use can enhance literature understanding~\cite{wang2025paperarena,pan2026paperscout}. Memory-augmented generation also shows that structured memory can improve downstream reasoning and personalized generation~\cite{tao2026memcast,yu2026memweaver}.

Tuning-based methods incorporating graph constraints further introduce structural priors into the generation process. By guiding decoding with subgraph-level plans or relation constraints, these approaches combine the fluency of PLMs with evidence-aware generation~\cite{li2024decoding,peng2024graph}.
Despite their substantial improvements in fluency, LLM-based methods remain prone to hallucination and weak evidence grounding, especially when summaries must faithfully reflect complex scientific evidence. This challenge highlights the need for mechanisms that not only enhance generation quality but also enforce explicit verification and logical coherence. Recent verification-oriented methods further reduce hallucinations by prompting models to generate, check, and revise their own outputs through explicit verification steps~\cite{dhuliawala-etal-2024-chain,gou2024critic}.

\section{Methodology}

In this section, we first define the problem, then present an overview of the proposed method, and finally describe the individual components in detail.

\subsection{Problem Definition}
Given a source paper $D$, the objective of scientific summarization is to generate a summary $S$ that preserves the document’s logical topology and factual details. To achieve this objective, ScholarSum formulates summarization as an iterative refinement process. Specifically, ScholarSum proceeds for at most $T$ iterations. At iteration $i$, the student produces a candidate summary $S_{\text{student}}^{(i)}$ by reasoning over a document-derived knowledge graph $G$, conditioned on the teacher’s structured feedback from the previous round, $F_T^{(i-1)}$. The teacher then evaluates $S_{\text{student}}^{(i)}$, returning a quality score $\sigma^{(i)}$ and updated feedback $F_T^{(i)}$. The process terminates when $\sigma^{(i)} \ge \theta$ or when the iteration budget $T$ is reached.

\begin{figure*}[t]
\centering
\includegraphics[width=\linewidth]{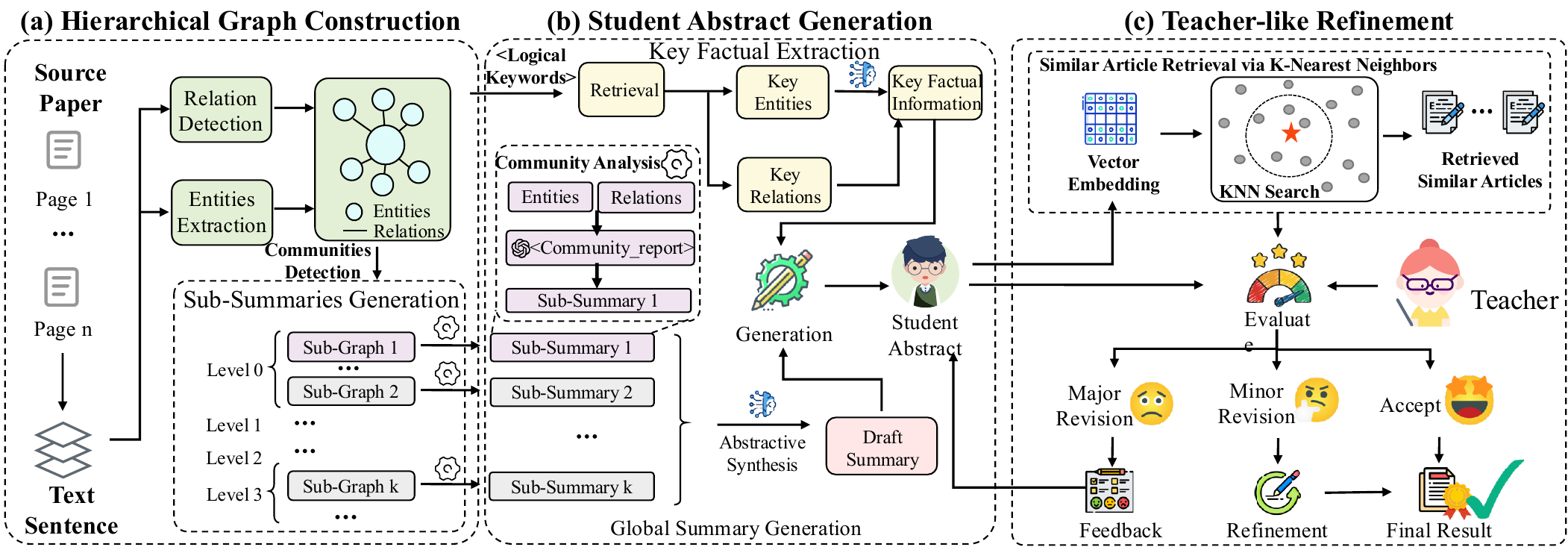}
\caption{Overview of ScholarSum for fluent and faithful scientific summarization: (a) hierarchical graph construction to model document-level structures, (b) student abstract generation to produce a logically coherent draft via structural guidance, and (c) teacher-like refinement to eliminate factual hallucinations through iterative self-correction.}
\label{fig:framework}

\end{figure*}

\subsection{Overview}
Figure~\ref{fig:framework} presents the ScholarSum framework for fluent and faithful scientific summarization. It proceeds through three main steps. First, hierarchical graph construction extracts entities and relations from source sentences to form subgraphs, capturing document-level structure. Second, student abstract generation produces coherent draft summaries guided by these subgraphs. Third, teacher-like refinement iteratively evaluates and revises the draft using retrieved similar articles, calibrating scientific writing style, expected content coverage, and discourse organization.

\subsection{Hierarchical Graph Construction}
The student drafts candidate summaries by operating over an explicit semantic workspace derived from the source paper. The key idea is to leverage document structure for planning and textual evidence for verification, ensuring that fluency and factuality reinforce each other rather than trade off.

To provide a manipulable representation of the paper's core claims, we construct a knowledge graph $G=(V,E)$ for a given paper $D$. We first split $D$ into sentence-level or paragraph-level textual units and use an instruction-tuned LLM with a constrained JSON schema to extract scientific entities $V$. Entities are typed into categories common in scientific writing, including tasks, methods, datasets, and metrics. For relation detection, we enumerate entity pairs within the same or adjacent textual units and classify them into typed relations, such as \textit{addresses}, \textit{uses}, \textit{evaluates-on}, \textit{reports}, and \textit{compares-with}. By externalizing the paper into this structured workspace, we enable both global planning and local evidence selection, providing a clear basis for coherent summary generation.
Scientific papers often contain multiple coherent threads, including problem motivation, methodological design, and empirical findings. To identify these thematic threads, we partition $G$ using the communities algorithm~\cite{traag2019leiden}:
\begin{equation}
\{G_1, G_2, \dots, G_k\} = \text{Leiden}(G),
\end{equation}
where each community $G_i$ corresponds to a focused theme. Tiny communities are removed and highly overlapping ones are merged to produce a stable plan that is neither fragmented nor overly coarse. This thematic planning ensures that the generated summaries respect the paper’s internal structure and maintain coherence across different sections.

\subsection{Student Abstract Generation}
Directly generating a scientific summary in a single pass often leads to either weak discourse organization or factual inaccuracies, as models lack an explicit mechanism to plan content and verify evidence jointly. To address this issue, the student treats the constructed graph not as a static representation but as an explicit semantic workspace that supports structured planning and evidence-aware generation.

Concretely, the student interacts with the graph through two complementary operations. Thematically, it navigates across communities to determine which themes to cover and in what order, using the community structure as a high-level outline where each $G_i$ corresponds to a candidate episode in the summary. Factually, it gathers evidence by querying salient triplets in the neighborhood of key nodes, such as proposed methods or evaluation settings, which serve as factual anchors for generation. Through this iterative interaction, the student repeatedly consults and updates its workspace, using structure to guide discourse flow and evidence to stabilize factual details.
Following this workspace-based view, the student generates summaries in two steps, reflecting a drafting–verification workflow. For each community $G_i$, the model produces a short sub-summary $s_i$ capturing the central contribution of that theme. These sub-summaries are then composed into a draft:
\[
S_{\text{draft}}^{(i)} = s_1 \oplus \dots \oplus s_k,
\]
where $\oplus$ denotes concatenation in thematic order. This draft provides a coverage-oriented scaffold that emphasizes completeness and logical organization. In the initial round, the draft is constructed without external feedback, while in later rounds, it can be lightly re-planned based on guidance from the teacher, such as reordering themes or expanding insufficiently covered content.

To improve factual fidelity, the student selects a compact set of key triplets from the graph to form a context subgraph $G_{\text{context}}^{(i)}$. This selection is guided by domain-specific cues that commonly signal high-value evidence in scientific writing, including experimental configurations, dataset identifiers, and main quantitative results. Conditioning on $G_{\text{context}}^{(i)}$ and the teacher's feedback, the student revises the draft as
\begin{equation}
S_{\text{student}}^{(i)} = \mathcal{F}(S_{\text{draft}}^{(i)}, G_{\text{context}}^{(i)}, F_T^{(i-1)}),
\end{equation}
with $F_T^{(0)}=\varnothing$. By grounding revision in explicit evidence, this step corrects issues identified during evaluation and produces a more fluent and factually reliable summary.

\subsection{Teacher-like Refinement}
Although the student is guided by structure and evidence, draft summaries may still contain omissions, logical inconsistencies, or subtle factual inaccuracies. This is because self-generated summaries lack an external reference for domain calibration and systematic quality control. To address this issue, we introduce a teacher-like refinement process that evaluates candidate summaries from a reviewer perspective and provides explicit signals for further improvement. 

To ground evaluation in the target scientific domain, the teacher retrieves $k$ similar papers $\{D_j^{\prime}: j=1,\dots,k\}$ and their abstracts $\{A_j^{\prime}: j=1,\dots,k\}$ from a reference corpus constructed from the training split of the corresponding benchmark, excluding the test paper itself. We encode the title and abstract of each paper using \texttt{text-embedding-3-small} and perform cosine-similarity KNN search over normalized embeddings. These retrieved abstracts serve as in-domain exemplars that help the teacher calibrate expected abstract structure, terminology usage, and the level of detail typically included in scientific summaries. The teacher then evaluates $S_{\text{student}}^{(i)}$ against these in-domain references and outputs both a scalar score and structured feedback:
\begin{equation}
\sigma^{(i)}, F_T^{(i)} = \mathcal{G}_{\text{evaluate}}(S_{\text{student}}^{(i)}, \{A_j^{\prime}: j=1,\dots,k\}),
\end{equation}
where $\sigma^{(i)}$ reflects overall quality, and $F_T^{(i)}$ records concrete revision targets, such as missing key contributions, incorrect factual statements, misordered logic, or unclear phrasing.

Based on $\sigma^{(i)}$, the Teacher selects one of three actions using thresholds $\theta_{\text{major}}$ and $\theta_{\text{minor}}$, where $\theta_{\text{minor}} \geq \theta_{\text{major}}$. 
When $\sigma^{(i)} \geq \theta_{\text{minor}}$, the summary is accepted and returned as the final output, denoted as \textit{Accept}. 
When $\theta_{\text{major}} < \sigma^{(i)} < \theta_{\text{minor}}$, 
the Teacher performs a light edit guided by $F_T^{(i)}$, denoted as \textit{Minor revision}:
\begin{equation}
S_{\text{final}} = \mathcal{F}_{\text{minor\_rev}}(S_{\text{student}}^{(i)}, F_T^{(i)}),
\end{equation}
where $\mathcal{F}_{\text{minor\_rev}}(\cdot)$ denotes the minor revision function, which lightly edits the candidate summary $S_{\text{student}}^{(i)}$ according to the Teacher's feedback $F_T^{(i)}$.
When $\sigma^{(i)} \leq \theta_{\text{major}}$, the Teacher returns $F_T^{(i)}$ to the Student as a conditioning signal for the next generation round, denoted as \textit{Major revision}.

By explicitly separating evaluation from generation and grounding feedback in retrieved in-domain references, this refinement process reduces hallucinations, improves factual fidelity, and provides a principled stopping criterion for iterative summarization. The interaction continues until the quality threshold is met or the iteration budget is reached.

\begin{table*}[t]
\centering
\small
\setlength{\tabcolsep}{4pt}

\begin{tabular*}{\textwidth}{@{\extracolsep{\fill}} lllccccc}
\toprule
\textbf{Dataset} & \textbf{Base LLM} & \textbf{Models} & \textbf{R-1} & \textbf{R-2} & \textbf{R-L} & \textbf{METEOR} & \textbf{BERTScore} \\
\midrule
\multirow{11}{*}{\textbf{ArXiv}} 
& \multirow{3}{*}{None} & T5 & 0.2638 & 0.0670 & 0.2323 & 0.1587 & 0.8273 \\
& & LED & 0.2267 & 0.0605 & 0.2000 & 0.1972 & 0.7739 \\
& & PEGASUS & 0.2550 & 0.0626 & 0.2034 & 0.1597 & 0.8131 \\
\cmidrule(l){2-8}
& \multirow{4}{*}{DeepSeek} & SumCot & 0.2027 & 0.0409 & 0.1837 & 0.2230 & 0.8128 \\
& & QA-prompting & 0.2635 & \underline{0.0694} & 0.2312 & 0.2362 & 0.8294 \\
& & Naive RAG & 0.2297 & 0.0377 & 0.1918 & 0.1759 & 0.7829 \\
& & \textbf{Ours} & \underline{0.2692} & \textbf{0.0708} & \underline{0.2362} & 0.2300 & \textbf{0.8360} \\
\cmidrule(l){2-8}
& \multirow{4}{*}{Qwen} & SumCot & 0.1940 & 0.0390 & 0.1730 & 0.2456 & 0.8133 \\
& & QA-prompting & 0.2339 & 0.0652 & 0.2097 & \underline{0.2539} & 0.8154 \\
& & Naive RAG & 0.2085 & 0.0330 & 0.1805 & 0.1832 & 0.7774 \\
& & \textbf{Ours} & \textbf{0.2764} & 0.0646 & \textbf{0.2412} & \textbf{0.2541} & \underline{0.8338} \\
\midrule
\multirow{11}{*}{\textbf{PubMed}} 
& \multirow{3}{*}{None} & T5 & 0.2560 & \underline{0.0809} & 0.2345 & 0.1427 & 0.8253 \\
& & LED & 0.2447 & 0.0739 & 0.2211 & 0.2100 & 0.7808 \\
& & PEGASUS & 0.2512 & 0.0687 & 0.2172 & 0.1364 & 0.8167 \\
\cmidrule(l){2-8}
& \multirow{4}{*}{DeepSeek} & SumCot & 0.1934 & 0.0299 & 0.1801 & 0.1818 & 0.8141 \\
& & QA-prompting & 0.2585 & 0.0663 & 0.2325 & 0.2187 & 0.8394 \\
& & Naive RAG & 0.2394 & 0.0485 & 0.2135 & 0.1790 & 0.7938 \\
& & \textbf{Ours} & \textbf{0.3102} & \textbf{0.0928} & \textbf{0.2834} & \textbf{0.2567} & \textbf{0.8531} \\
\cmidrule(l){2-8}
& \multirow{4}{*}{Qwen} & SumCot & 0.2060 & 0.0412 & 0.1851 & 0.2312 & 0.8191 \\
& & QA-prompting & 0.2634 & 0.0748 & 0.2410 & 0.2496 & 0.8316 \\
& & Naive RAG & 0.2382 & 0.0511 & 0.2114 & 0.2029 & 0.7931 \\
& & \textbf{Ours} & \underline{0.2929} & 0.0735 & \underline{0.2645} & \underline{0.2505} & \underline{0.8435} \\
\bottomrule
\end{tabular*}

\caption{Main experimental results on the ArXiv and PubMed datasets. Best results are highlighted in \textbf{bold}, and second-best are \underline{underlined}.}
\label{tab:main_results}
\end{table*}
\section{Experiments}

In this section, we evaluate the effectiveness and factual reliability of ScholarSum. Experiments are conducted on multiple standard scientific summarization benchmarks, with comparisons against strong baselines.

\subsection{Experimental Settings}
\paragraph{Datasets and Metrics.} 
We evaluate ScholarSum on multiple scientific summarization benchmarks: ArXiv and PubMed~\cite{cohan-etal-2018-discourse}. For evaluation, we report ROUGE (R-1, R-2, R-L)~\cite{lin2004rouge}, METEOR~\cite{banerjee2005meteor}, and BERTScore~\cite{zhang2019bertscore} to assess lexical coverage, semantic matching, and contextual similarity. ROUGE emphasizes lexical overlap, METEOR captures semantic matching with stemming and synonymy, and BERTScore measures contextual alignment via encoders.

\paragraph{Baselines.} We compare against three groups of competitive baselines: 
(1) Traditional encoder--decoder summarization models, including T5~\cite{raffel2020exploring}, LED~\cite{beltagy2020longformer}, and PEGASUS~\cite{zhang2020pegasus}; 
(2) LLM-based prompting methods, namely SumCoT~\cite{wang2023element} and QA-prompting~\cite{sinha2025qa}, instantiated with both DeepSeek and Qwen base models; 
(3) Naive RAG~\cite{lewis2020retrieval}, a retrieval-augmented LLM method. This baseline is particularly important because ScholarSum also uses retrieval; thus, it isolates the benefit of structured graph reasoning and teacher-guided refinement beyond adding retrieval alone.
To ensure a fair comparison, all LLM-based baselines share the same backbone models as ScholarSum (i.e., DeepSeek and Qwen). We also enforce consistent decoding constraints, such as temperature and output length budget, across all methods whenever applicable. 

\subsection{Main Results}
Table~\ref{tab:main_results} comprehensively compares summarization performance across the ArXiv and PubMed benchmark datasets. ScholarSum demonstrates strong performance in most scenarios, often achieving the best or second-best scores across both lexical (ROUGE) and semantic (BERTScore) metrics. Notably, LLM-based baselines like QA-prompting also show competitive results, particularly in generating linguistically fluent text. This further validates the potential of leveraging large language models for scientific abstract writing. However, these models often lack the explicit structural guidance and self-correction mechanisms introduced by our framework, limiting their ability to preserve both content coverage and factual consistency. Retrieval-based baselines such as Naive RAG perform moderately in capturing local evidence, but their performance drops regarding structural coherence (R-L) due to the fragmentation of retrieved chunks.

This contrast underscores the importance of balancing global logical planning with fine-grained evidence grounding. Beyond the overall metrics, the performance gap is particularly pronounced in the PubMed dataset, where scientific terminology and dense experimental data are more prevalent. While standard generative methods often struggle with the domain-specific complexity of medical literature, ScholarSum maintains high precision through its iterative verification loop. This suggests that our approach is especially effective for documents requiring a high degree of technical accuracy. In summary, the consistent gains can be attributed to the core design of ScholarSum: constructing a hierarchical reflective graph to guide a student–teacher iterative process. This unified approach effectively reconciles the inherent conflict between linguistic fluency and factual faithfulness, addressing the critical challenges and structural consistency requirements in real-world scientific summarization.

\begin{figure}[t] 
    \centering
    \includegraphics[width=\linewidth]{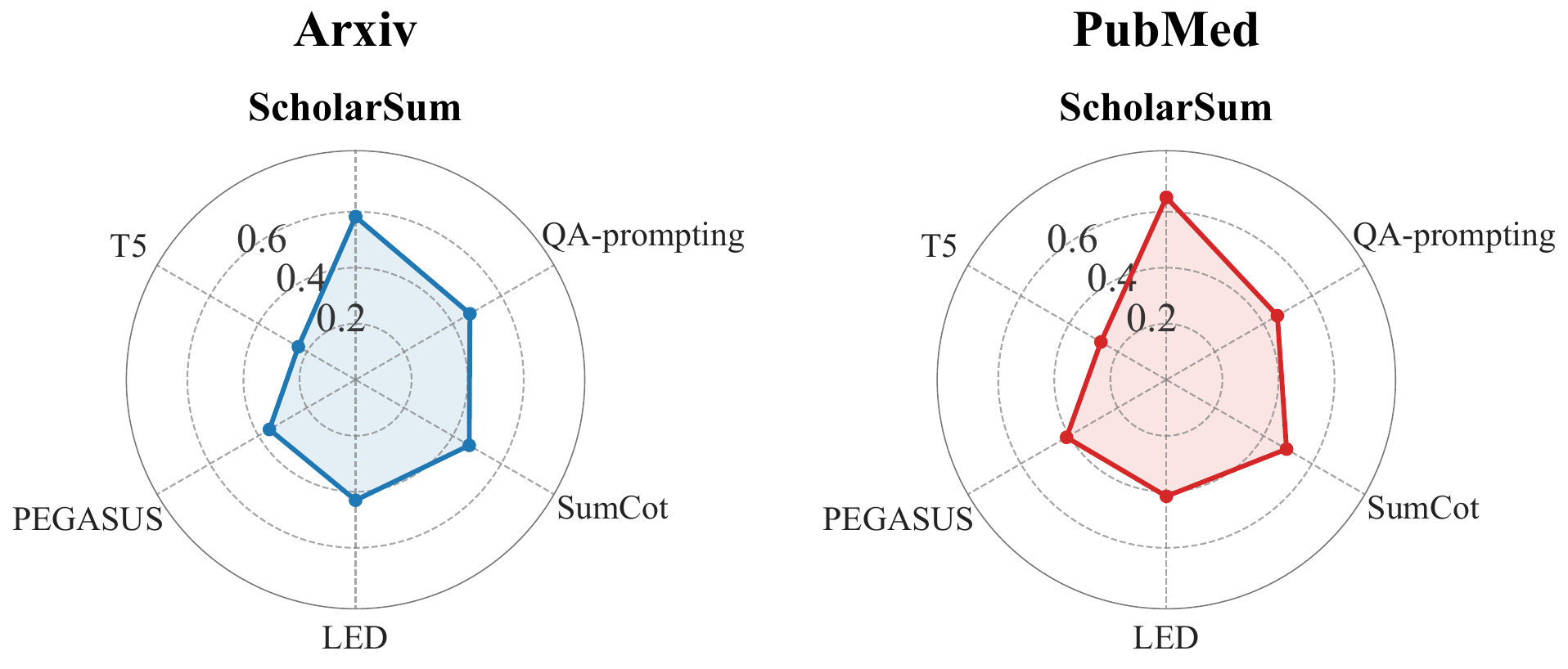}
    \caption{Evaluating the factual consistency of ScholarSum across diverse scientific datasets.}
    \label{fig:factuality_results}
\end{figure}

\subsection{Factual Consistency Analysis}
\paragraph{Factuality Metric.}
We evaluate factual consistency using MiniCheck~\cite{tang-etal-2024-minicheck}, which estimates whether the generated summary is supported by the source document. To discourage trivially short but factually safe summaries, we introduce a brevity penalty. Let $c$ and $r$ denote the token lengths of the generated summary and the reference abstract, respectively. The brevity penalty is defined as:
\begin{equation}
BP =
\begin{cases}
1, & c \ge r, \\
\exp(1-r/c), & c < r.
\end{cases}
\end{equation}
We then compute the final factuality score as:
\begin{equation}
S_{\text{fact}} = S_{\text{MiniCheck}} \times BP,
\end{equation}
where $S_{\text{MiniCheck}}$ is the MiniCheck score between the generated summary and the source document. A higher $S_{\text{fact}}$ indicates that the summary is better supported by the source document while maintaining sufficient content coverage.

Figure~\ref{fig:factuality_results} analyzes how different architectural components contribute to the factual reliability of generated summaries. Incorporating the hierarchical reflective framework leads to substantial improvements, whereas variants using standard prompting or vanilla retrieval consistently underperform.
We further examine two key components: hierarchical graph-based structure construction, which supports macro-level thematic organization and global structural guidance, and the refinement mechanism, which enables iterative claim-level verification and correction. While both components improve factual consistency, the refinement mechanism typically yields larger gains in reducing hallucinations.
The strong performance of the full model, particularly its clear advantages on ArXiv and PubMed, demonstrates its ability to integrate global structural context with fine-grained evidentiary verification, thereby enhancing summary faithfulness without compromising content richness.

\begin{figure}
    \centering
    \includegraphics[width=1\linewidth]{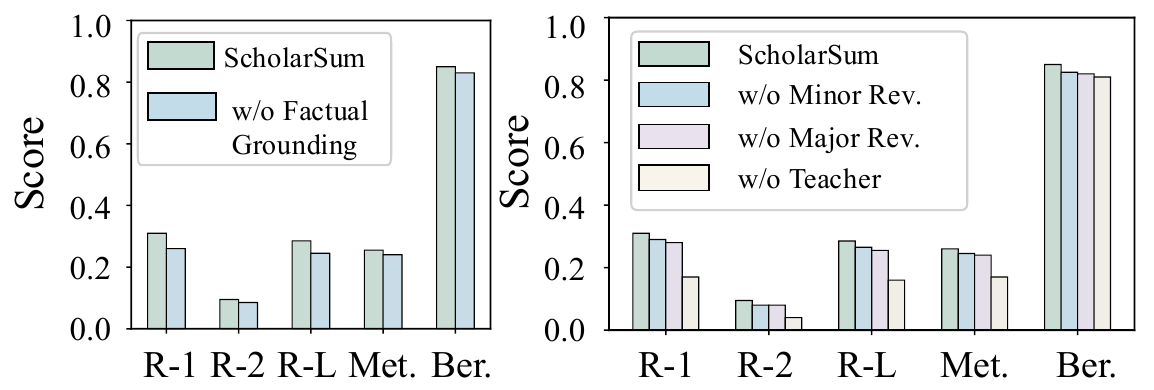}
    \caption{Analysis of component contributions via ablation studies on the student (left) and the teacher (right) modules.}
    \label{fig:ablation_studies}
\end{figure}
\begin{figure}[t]
  \centering
  \includegraphics[width=1\linewidth]{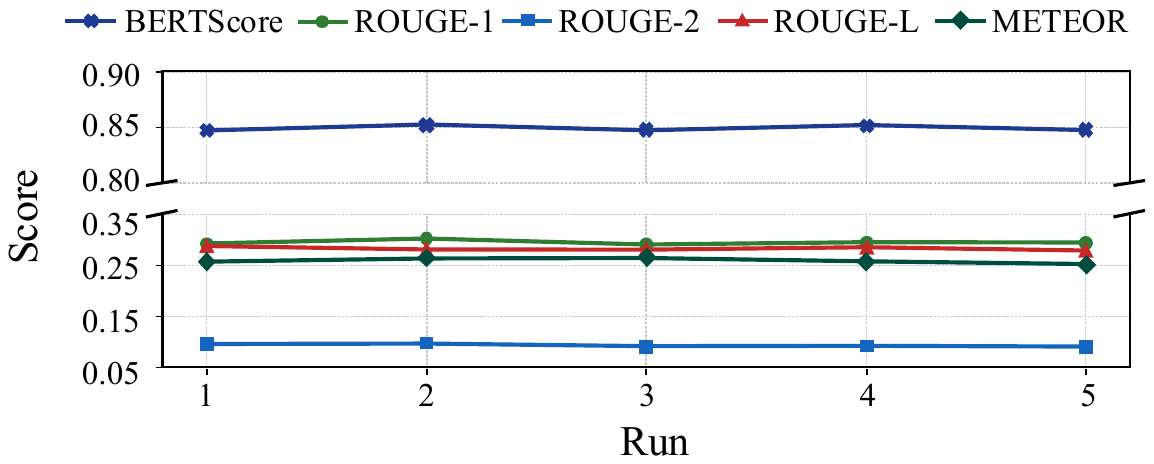}
  \caption{Comprehensive analysis of model performance stability across multiple datasets.}
  \label{fig:stability}
\end{figure}
\subsection{Ablation Analysis}
To assess the contributions of key design choices, we perform systematic ablations on both the Student and Teacher. The results show that factual grounding is necessary for faithful detail retention, while teacher-guided revision is essential for correcting content selection and discourse organization errors that persist in one-shot generation.

\paragraph{Ablation on the Student Module.} 
Figure~\ref{fig:ablation_studies} (left) examines how factual extractive grounding influences generation performance. The variant without factual grounding consistently underperforms across all metrics, suggesting that generation without explicit evidence support is less reliable. In contrast, incorporating extracted factual evidence provides stable and significant gains, underscoring the importance of evidence-based anchoring for factual correctness. The superior performance of the full model indicates its ability to effectively integrate extracted evidence into the generation process, thereby enhancing factual accuracy.

\paragraph{Ablation on the Teacher Module.} 
Figure~\ref{fig:ablation_studies} (right) analyzes the contribution of the teacher module to overall performance. Removing the teacher leads to the largest performance drop among all ablation settings, indicating that reflective revision plays a central role in improving abstract quality. We further examine different revision pathways within the teacher. The major revision pathway yields more substantial gains than minor edits, suggesting that high-level feedback on content selection and discourse organization is more critical for scientific abstracts. In particular, omissions of core contributions or misplacement of key results tend to degrade overall quality more severely than surface-level wording issues.

\begin{figure*}[t!]
\centering
\includegraphics[width=0.92\textwidth]{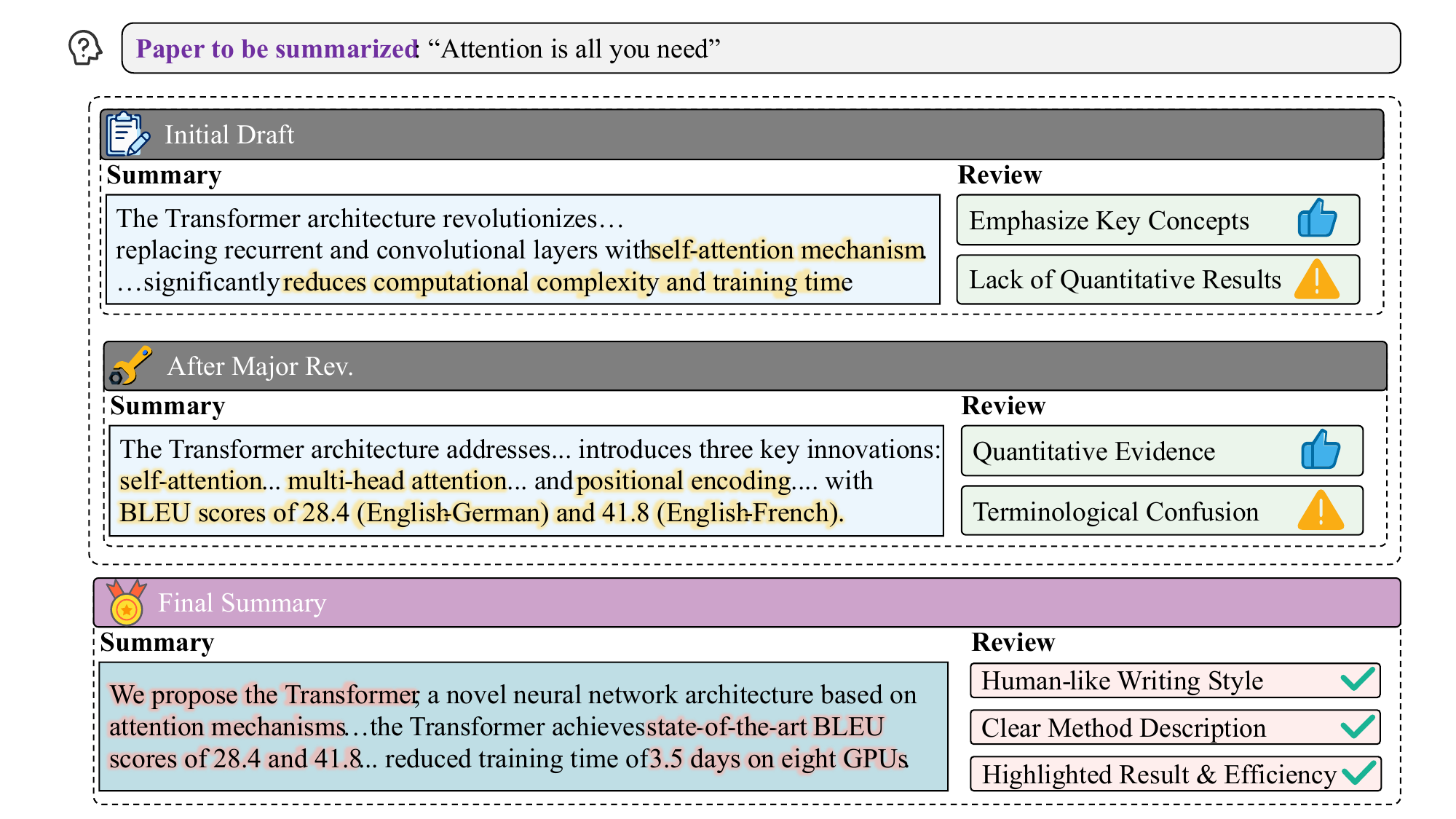}
\caption{
Case study of ScholarSum's iterative refinement on the ``Attention Is All You Need'' abstract, illustrating the incremental evolution of the resultant high-fidelity summary.
}
 \label{fig:case_study}
\end{figure*}

\begin{table}[t]
\centering
\resizebox{\columnwidth}{!}{%
\begin{tabular}{lccccc}
\toprule
\textbf{Temperature} & \textbf{R-1} & \textbf{R-2} & \textbf{R-L} & \textbf{METEOR} & \textbf{BS} \\
\midrule
1.0 & 0.2880 & 0.0806 & 0.2650 & 0.2472 & 0.8484 \\
\textbf{0.8} & \textbf{0.3102} & \textbf{0.0928} & \textbf{0.2834} & \textbf{0.2567} & \textbf{0.8531} \\
1.3 & \underline{0.2869} & \underline{0.0801} & \underline{0.2614} & \underline{0.2421} & \underline{0.8487} \\
0.2 & 0.2789 & 0.0713 & 0.2550 & 0.2262 & 0.8452 \\
\bottomrule
\end{tabular}
}
\caption{Impact of decoding temperature on generation quality.}
\label{tab:hyper_temp}
\end{table}

\subsection{Exploration Analysis}
\paragraph{Stability Analysis.}
Figure~\ref{fig:stability} shows the performance of ScholarSum across five independent runs with different random seeds. Overall, the results demonstrate a high degree of consistency on all evaluation metrics. The standard deviations of ROUGE scores remain very small (e.g., approximately $0.004$ for ROUGE-L), and similarly low variance is observed for METEOR and BERTScore, indicating that the framework is not sensitive to random initialization. Across runs, performance fluctuations are minimal, and no metric exhibits notable instability. This behavior can be attributed to the structured workflow of ScholarSum. The semantic workspace constrains global planning and content organization, while the teacher provides corrective feedback during reflective revision, together reducing the impact of stochastic decoding noise and leading to reliable generation quality.

\begin{table}[t]
\centering
\resizebox{\columnwidth}{!}{%
\begin{tabular}{lccccc}
\toprule
\textbf{Keyword Ver.} & \textbf{R-1} & \textbf{R-2} & \textbf{R-L} & \textbf{METEOR} & \textbf{BS} \\
\midrule
task & 0.2835 & 0.0804 & 0.2545 & 0.2446 & 0.8469 \\
\textbf{query1} & \underline{0.3093} & 0.1048 & \textbf{0.2929} & \textbf{0.2575} & \textbf{0.8528} \\
query2 & 0.3072 & \textbf{0.1082} & 0.2840 & 0.2515 & 0.8501 \\
query3 & 0.3014 & 0.0978 & 0.2883 & 0.2473 & 0.8515 \\
LLM-adaptive & \textbf{0.3098} & \underline{0.0979} & \underline{0.2909} & \underline{0.2529} & \underline{0.8454} \\
\bottomrule
\end{tabular}
}
\caption{Analyzing different logical keyword selection strategies.}
\label{tab:hyper_keys}
\end{table}

\paragraph{Hyperparameter Sensitivity Analysis.}
Table~\ref{tab:hyper_temp} analyzes how two key hyperparameters—decoding temperature and logical keyword selection strategy—affect generation performance. Adjusting temperature reveals a trade-off: lower values make the model overly conservative, limiting diversity, while higher values introduce randomness that reduces coherence and increases factual errors. The results show that a temperature of 0.8 best balances diversity and factual reliability. Different keyword selection strategies also influence content fidelity and relevance, highlighting the importance of hyperparameter tuning. The performance under settings demonstrates the model’s ability to generate coherent, factually grounded text while maintaining sufficient variability.

Table~\ref{tab:hyper_keys} presents a comparative analysis of keyword selection strategies for evidence grounding. Overall, explicit query constraints clearly outperform unguided generation, indicating that structured retrieval guidance is important for anchoring high-value facts, such as dataset names and numerical results, and thus helps reduce hallucinations. We also observe a trade-off between precision and coverage across different query variants, as reflected by the variations in ROUGE-2 and ROUGE-L. Notably, the LLM-adaptive strategy achieves performance comparable to manually designed queries. Although it does not consistently achieve the best results across all metrics, it reduces the dependence on hand-crafted query templates, suggesting the potential of more automated evidence selection under realistic deployment scenarios.

\subsection{Case Study Analysis}
To qualitatively illustrate our iterative refinement process, we present a case study on the influential paper ``Attention Is All You Need''~\cite{vaswani2017attention}. Figure~\ref{fig:case_study} shows how the generated summary evolves across multiple revision stages.
The initial draft captures the core idea of replacing recurrent and convolutional architectures with self-attention, but it lacks quantitative evidence and a clear argumentative structure. Guided by the teacher module, the major revision reorganizes the summary into a problem-solution narrative and incorporates key empirical results, such as BLEU scores and training efficiency. The final revision further improves conciseness, terminology precision, and abstract-style writing. This progression shows that ScholarSum can address common summarization errors, including missing numerical evidence, terminology drift, and weak discourse organization, through teacher-guided iterative refinement.

\section{Conclusion}
In this work, we proposed ScholarSum, a hierarchical reflective framework for faithful and fluent scientific summarization. ScholarSum organizes source documents into hierarchical knowledge graphs to capture global logic and guide structured draft generation, and then employs a teacher-like refinement process to detect unsupported content and trigger targeted revision. This student--teacher design combines macro-level discourse planning with fine-grained evidence grounding, thereby improving both content completeness and factual faithfulness. Experiments on ArXiv and PubMed show that ScholarSum consistently outperforms strong baselines, while ablation and stability analyses further validate the effectiveness and robustness of its key components. Future work will extend ScholarSum to broader scientific domains and incorporate multimodal evidence from figures and tables.
\section*{Acknowledgments}

This research was supported by grants from the National Natural Science Foundation of China (No. 62502486), the Provincial Natural Science Foundation of Anhui Province (No. 2408085QF193), USTC Research Funds of the Double First-Class Initiative (No. YD2150002501), and the Fundamental Research Funds for the Central Universities of China (No. WK2150110032).

\section*{Contribution Statement}

Bohou Zhang and Xiaoyu Tao contributed equally to this work. Mingyue Cheng supervised the project, provided guidance throughout the study, and served as the corresponding author. All authors reviewed the final manuscript.

\bibliographystyle{named}
\bibliography{ijcai26}

\end{document}